\documentclass[runningheads]{llncs}
\usepackage[ruled,vlined]{algorithm2e}
\usepackage{graphicx}
\usepackage{amsmath}
\usepackage{theorem}   %for mathematical theorems and lemmas
\usepackage{adjustbox}
\usepackage{multirow}
\usepackage{longtable}
\usepackage[protrusion=true,expansion=true]{microtype}
\usepackage[bitstream-charter]{mathdesign}
\usepackage[T1]{fontenc}
\usepackage{enumitem}
\usepackage{array}
\usepackage{comment}

\usepackage[english]{babel}
\usepackage[nottoc]{tocbibind}

\begin{document}
\title{RATAS: A Generative AI Framework for Explainable and Scalable Automated Answer Grading} 

\author{Masoud Safilian\inst{1} \and
Amin Beheshti\inst{1}\and Stephen Elbourn \inst{2}}
\authorrunning{M. Safilian and A. Beheshti}
\institute{Macquarie University, Sydney, Australia \\
\email{masoud.safilian@hdr.mq.edu.au} \\
\email{amin.beheshti@mq.edu.au}
\and ITIC Company \\
\email{stephen.elbourn@itic.com.au}
}

%
\maketitle              % typeset the header of the contribution
\begin{abstract}
Automated answer grading is a critical challenge in educational technology, with the potential to streamline assessment processes, ensure grading consistency, and provide timely feedback to students. However, existing approaches are often constrained to specific exam formats, lack interpretability in score assignment, and struggle with real-world applicability across diverse subjects and assessment types. To address these limitations, we introduce RATAS (Rubric Automated Tree-based Answer Scoring), a novel framework that leverages state-of-the-art generative AI models for rubric-based grading of textual responses. RATAS is designed to support a wide range of grading rubrics, enable subject-agnostic evaluation, and generate structured, explainable rationales for assigned scores. We formalize the automatic grading task through a mathematical framework tailored to rubric-based assessment and present an architecture capable of handling complex, real-world exam structures. To rigorously evaluate our approach, we construct a unique, contextualized dataset derived from real-world project-based courses, encompassing diverse response formats and varying levels of complexity. Empirical results demonstrate that RATAS achieves high reliability and accuracy in automated grading while providing interpretable feedback that enhances transparency for both students and instructors. To advance research in this domain, we publicly release all code on https://github.com/datalab912/RATASv1.

\end{abstract}

\section{Introduction}

Open-ended (OE) questions are integral to education, fostering critical thinking, creativity \cite{openended}, and self-explanation \cite{selfexplanation}. Rubrics play a crucial role in ensuring fair and structured assessment of OE's based exams, providing consistency in grading \cite{kolappan2023computer}, and providing constructive feedback \cite{santos2009enhancing}. However, OE manual grading remains labor-intensive \cite{weegar2024reducing}, error-prone, and inefficient, especially for large-scale assessments. 

Most research on automated grading focuses on short-answer grading, essay scoring, and writing evaluation \cite{gao2024automatic}, specifically targeting key narrative components of real-world exams. Automated Short Answer Grading (ASAG) methods are typically restricted to responses ranging from a single phrase to a short paragraph \cite{burrows2015eras}, whereas the recently introduced Automatic Long Answer Grading (ALAG) has been evaluated on responses averaging 120 words \cite{sonkar2024automated}. Existing approaches face several challenges, including inadequate rubric integration \cite{prasain2020analysis}, as most methods treat rubrics as unstructured lists of criteria \cite{wang2019inject}; limited generalizability across different exam subjects, requiring retraining for each new exam and struggling with complex grading structures; and a lack of explainability \cite{condor2024explainable}. 

In this paper, we formally define and mathematically formulate Rubric-Based Automatic Grading for Textual Exams (RAGT), codifying real-world rubric structures and scoring logic. RAGT presents a highly complex Natural Language Processing (NLP) challenge due to the diversity of exam types, the complexity of real-world rubrics, the open-ended nature of textual responses, and the scarcity of labeled data. Additionally, it requires generating structured, feedback-oriented rationales, further increasing its complexity.

To address the limitations of existing methods, we introduce RATAS (Rubric Automated Tree-based Answer Scoring), a novel framework leveraging state-of-the-art generative AI to enable accurate, subject-agnostic, and interpretable rubric-based grading. Over the past two decades, artificial intelligence (AI) has evolved from traditional methods like combinatorial optimization \cite{saf1} and sampling-based planning \cite{saf2} to ML, NLP, and more recently, transformer architectures \cite{attention} and generative AI \cite{GenAI}, which has gained significant traction in education \cite{EDUGAI}.  Recent advances in foundation, transformer-based generative AI, and large language models (LLM) \cite{LLM}, despite their success in NLP tasks, have demonstrated suboptimal performance in ASAG, highlighting the need for a more structured approach to rubric-based automated grading \cite{LLMlimitation}.

RATAS systematically contextualizes rubrics by generating a rubric-based tree, enhancing both answer analysis and scoring reliability. It decomposes the grading process into modular Downstream NLP Tasks (DNTs), enabling strong performance on longer answers, improving reliability, and providing structured feedback and reasoning. Additionally, it enhances adaptability across diverse exam formats while reducing complexity and minimizing training data requirements. RATAS also allows instructors to combine rubric criteria, streamlining rubric design. Achieving these objectives necessitates the design and development of effective DNTs tailored for RATG, integrating them into a robust architecture, properly adapting LLMs for each DNT, and implementing a flexible rubric-based scoring mechanism.

Due to the lack of datasets for RATG, we evaluate RATAS using a real-world dataset we created from university-level project-based courses, consisting of longer answers than existing datasets. By implementing RATAS architecture and employing GPT-4o \cite{GPT4o} for addressing DNTs in its lower layers, we compare its performance against direct GPT-4o usage and manual grading. Results show that RATAS significantly improves grading accuracy, interpretability, and scalability, reinforcing its potential as a robust solution for automated rubric-based assessment.

\section{Related Work}

    Most research on automated grading has focused on ASAG, developed and evaluated on datasets with an average response length of fewer than 20 words, such as Beetle \cite{dzikovska2016joint} and Texas 2011 \cite{mohler2011learning}. One notable exception is \cite{sonkar2024automated}, which considers longer answer grading on the RiceChem dataset (average answer length: 120 words) by framing grading as a textual entailment task, achieving approximately 65\% accuracy on four questions. Among ASAG methods, neural network (NN) and deep learning approaches have shown strong performance, including siamese networks \cite{yeruva2022triplet} and LSTM-based models \cite{erickson2020automated,tulu2021automatic}. 

Building on these NN methods, LLMs have recently garnered attention for their extensive knowledge bases, contextual understanding, and adaptability \cite{latif2024fine}. For instance, \cite{balaha2021automatic} employed a semantic similarity approach by training a USE, attaining 77.95\% accuracy. Meanwhile, \cite{zhu2022automatic} transformed the grading task into a multi-class classification problem, fine-tuning a BERT encoder-based model \cite{devlin2018bert} for scoring. With the emergence of GPT-based models \cite{kublik2023gpt}, researchers have begun comparing their performance to more traditional architectures. In 2024, \cite{latif2024fine} demonstrated that a fine-tuned GPT-3.5 model achieved an average accuracy of 91.5\%, surpassing a fine-tuned BERT model’s 83.8\%. Similarly, GPT-4 has been shown to outperform SVM and match or exceed other strong baselines, albeit with some overfitting in GPT-3.5 \cite{chamieh2024llms}. Another study reported a 0.677 quadratic weighted kappa for GPT-4 across ten questions, although its performance varied by subject \cite{jiang2024short}. 

Generating meaningful feedback remains a challenge in automated grading. Some approaches use NLP-based systems for diagnostic scores and argumentation suggestions \cite{lee2019automated}, while some employ simulation-based feedback to enhance learning \cite{lee2021machine} Recent efforts focus on explainable AI (XAI), integrating neural additive models \cite{condor2024explainable} and other methods \cite{singh2023explaining} to improve grading transparency and interpretability.

\section{RAGT Problem Definition}
Reliable automatic scoring in RATG requires a comprehensive, well-structured rubric covering all designated rules. We first define the RATG paradigm and develop a scoring logic to simplify rubric creation for instructors. RATAS is implemented based on this definition to \textbf{handle a wide range of extensive real-world rubrics}. 

\textbf{Rubric structure}: Rubrics lie at the heart of this problem, containing the rules and policies necessary for grading \cite{el2022using}. They commonly take one of two forms: holistic rubrics, providing an overarching perspective on performance, and analytical rubrics, offering detailed criteria and scoring levels that facilitate more reliable automatic assessments \cite{wiseman2012comparison}. Table \ref{tab1} presents our redefined rubric structure—an analytical approach incorporating all essential scoring components. Based on the rubrics required components \cite{dawson2017assessment} and our evaluation of multiple course rubrics, most existing rubrics can be readily adapted to this format, ensuring comprehensive coverage in a range of educational contexts. 

Typically, the second rubric attribute (\textit{Basic-rule}) serves as a criterion title, with corresponding details reflected in \textit{Level-of-Achievement (la)}. In certain cases, such as the real rubric used in this paper evaluation, rules may instead appear within the \textit{Basic-rule} itself (as seen in the second example of Tabl \ref{tab1}), leaving \textit{la} empty or providing additional scoring details as needed.

\begin{table}
\caption{Rubric columns definition, The essential column is indicated with a star.}\label{tab1}
    \begin{tabular}{|p{1.8cm}|p{2.9cm}|p{7.5cm}|} \hline 
         \textbf{Column}& \textbf{Column Description} &\textbf{Example}\\ \hline 
         ID: \textit{Integer}*& Show the unique ID for each line of rule &1\\ \hline 
         Basic-Rule: \textit{string}*& A set of words that may serve as a criterion title or encompass some set of rules &Example 1: BERT Architecture and Applications
Example 2: The response must explain the BERT model's architecture and describe at least one of its applications in NLP tasks.\\ \hline 
         Score-Source (SS) : \textit{percentage}*& Represents the exam score percentage if all criteria are met&SS = 30\%\\ \hline 
         Level-of-Achievement: \textit{List of string and their related percentage}& For each rule, a list of levels of achievement (la), including level of quality (lq) and corresponding scores (levels of score, ls) can represent varying performance levels. &\textbf{la1:} (\textbf{lq 1:} The response provides a clear and accurate explanation of BERT's layers and attention mechanism. It correctly identifies and thoroughly explains at least one NLP application of BERT, including fine-tuning methods for each task., \textbf{ls 1:} 100\%) \newline
\textbf{la2}: (\textbf{lq2}: The response describes BERT's layers and attention mechanism but lacks key components or contains inaccuracies. It identifies an NLP application and fine-tuning methods but lacks clarity or detail., \textbf{ls2}: 50\%)\\ \hline
    \end{tabular}
\end{table}
\textbf{Logic of Rubric-based Scoring:} In typical rubric-based scoring, an instructor or system evaluates each criterion (defined in the \textit{Basic-Rule}, if it is not a title) to determine whether the requirement is met. If the criterion is met, the next step is to assess which lq, as defined in the related \textit{la} for that row, is more satisfied by the answer. The score for each row is then assigned by multiplying the \textit{ls} associated with the matched quality level by the \textit{ss} percentage. 

Achieving reliable automatic means listing each criterion on its own row, along with detailed \textit{la} definitions that cover all possible lq. To address the challenge of the difficulty of designing complete rubrics, we propose a more \textbf{flexible scoring\textbf{ }logic} that reduces the \textbf{complexity of rubric design} while preserving its core principles. Under this approach, educators can rely on fewer rows and simpler \textit{la} lists without compromising accuracy—provided that the underlying scoring framework supports both these streamlined rubrics and more typical, fully detailed formats.  

\textbf{Flexible Rubric-Based Scoring Logic:} In this approach, the system first determines the \textit{Score Percentage (SP)}—indicating how thoroughly the answer meets each row’s \textit{Basic-Rule}—rather than simply checking for a complete match. Next, it identifies the most suitable \textit{la} and computes a \textit{Level of Quality Aligning Percentage (LQAP)}, reflecting how closely the answer meets the chosen quality level. The final score is calculated based on \(Achieved Score=SP×LQAP×related\_ls×related\_SS\), where \textit{related\_SS} is the row’s score source and \textit{related\_ls} is the score level associated with the  maximum matched \textit{la}. 

In practice, educational rubrics are often highly intricate, featuring multiple rules that can make a fully detailed rubric using typical logic overly complex or even impractical. Our flexible scoring logic \textbf{does not alter the nature or definition of RATG, nor does it invalidate the use of typical rubrics. }Instead, it streamlines rubric design for systems that support this method. Crucially, any rubric formulated with this logic can be replicated via traditional approaches, preserving the same rules and scoring outcomes. For example, the third column of Table \ref{tab1} illustrates a single row with two \textit{\textbf{la}}s; in a purely typical rubric, this would require two rows and four \textit{\textbf{la}}s, a difference that grows with more complex rules.

\textbf{Mathematical Definition of Basic ARSN Problem}: Given an answer \( A \) and a rubric\( R \), the objective of RATG is to compute a score \( S \) , ranges from 0 to the maximum score \( MS \), that reflects how closely \( A \) adheres to \( R \): \( S = \arg \max_{s \in [0, MS]} P(S = s \mid A, R)
\). As summarized in Table~\ref{tab1}, the rubric has \(n\) rows. Let \(S_i\) denote the score awarded for the \(i\)-th row, with \(r_i\) the corresponding \emph{Basic-Rule}, \(ss_i\) the \emph{Score-Source}, and \(la_i\) the associated \textit{level-of-Achivement}. The total score \(S\) is:
\[
S \;=\; \sum_{i=1}^{n} S_i 
\;=\; \sum_{i=1}^{n} \arg\max_{s \in [0,\,ss_i]} P\bigl(S_i = s \,\mid\, r_i,\, la_i\bigr).
\]

Each \(la_i\) consists of \(m\) pairs \(\{(lq_{ij},\; ls_{ij})\}\), where \(lq_{ij}\) is the \(j\)-th \emph{Level of Quality} and \(ls_{ij}\) the corresponding score. To incorporate a flexible scoring logic, define \(SP_i\) (score percentage) as the proportion of the rule \(r_i\) satisfied in \( A \) and \(LQAP_{ij}\) (Level-of-Quality Matching Percentage) as the proportion of \(lq_{ij}\) matched by \(as_i\). The refined score formula for \(S\) becomes:
\[
S 
= \sum_{i=1}^{n} S_i 
= \sum_{i=1}^{n}
\Bigl(SP_i \,\times\, LQAP_{i,\max} \,\times\, LS_{i,\max} \,\times\, ss_i\Bigr), \qquad where
\]
\[
SP_i 
= \arg\max_{sp \in [0,1]} 
P\bigl(SP = sp \,\mid\,  r_i\bigr),
\]
\[
LQAP_{i,\max} 
= \max_{j \in \{1,\dots,m\}}\!\bigl(LQAP_{ij}\bigr), 
\qquad
LQAP_{ij} 
= \arg\max_{lqap \in [0,1]} 
P\bigl(LQAP = lqap \,\mid\, as_i,\; lq_{ij}\bigr),
\]
and \(LS_{i,\max}\) is the score \(ls_{ij}\) corresponding to the maximal \(LQAP_{ij}\).

\section{RATAS Processes, Conceptual Design and Model Architecture}
\begin{figure}
\includegraphics[width=\textwidth]{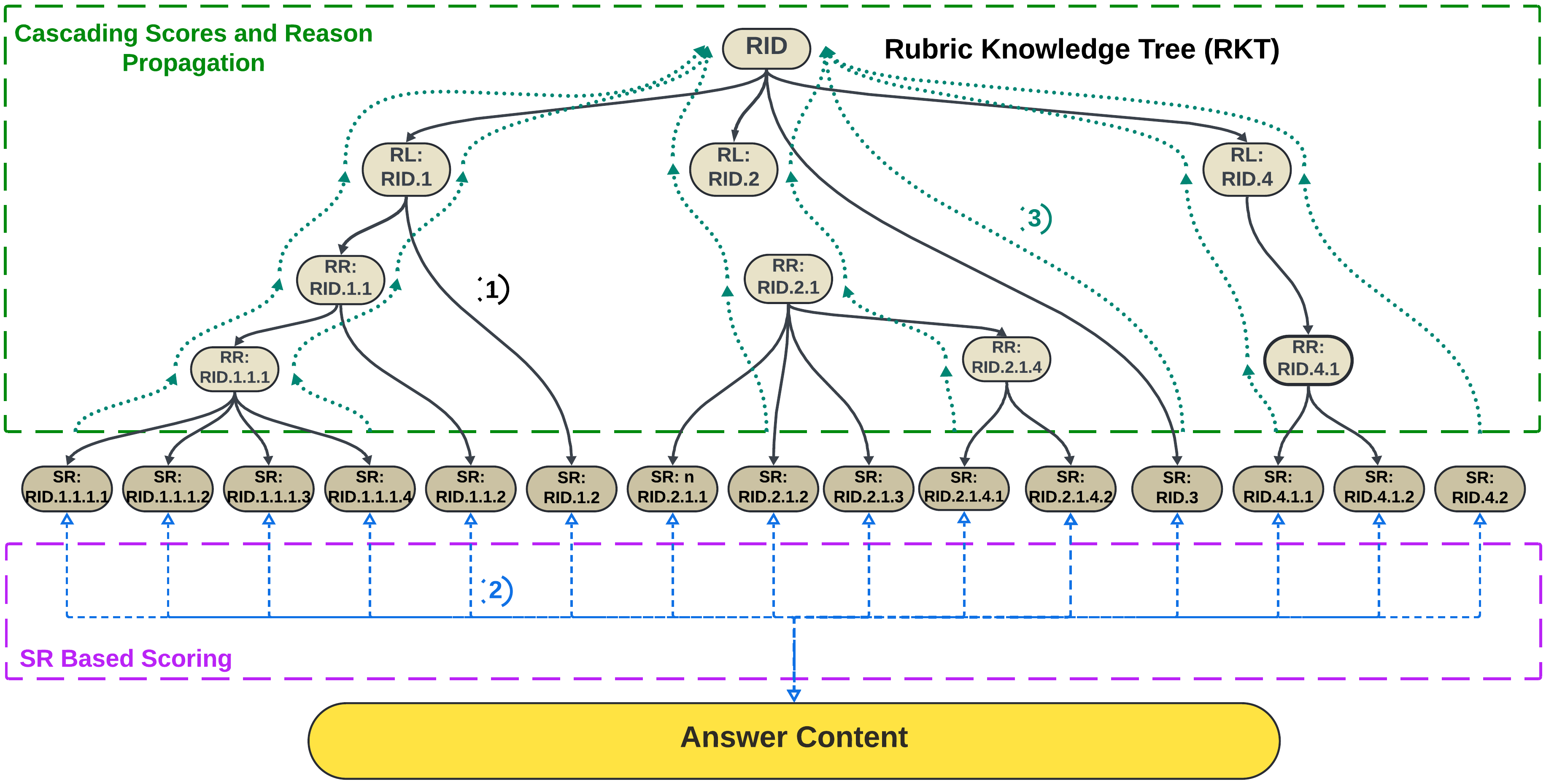}
\caption{The RATAS process flow for RKT construction and scoring involves: 1) Constructing RKT nodes from the root. 2) Scoring and reasoning based on SRs, updating SR node attributes. 3) Cascading scores and reason propagation, updating internal RKT node attributes
  } \label{fig1}
\end{figure}

\begin{figure}
\includegraphics[width=\textwidth]{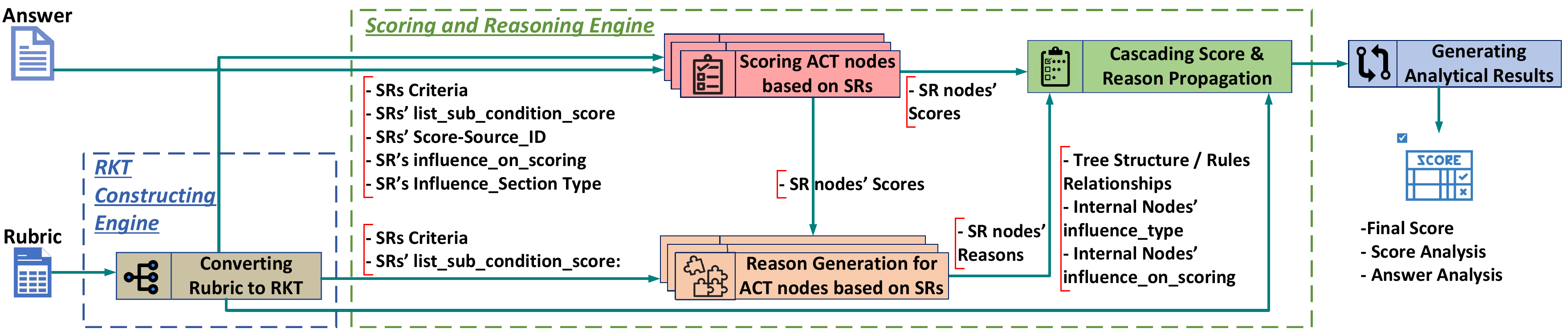}
\caption{RATAS Conceptual Steps and Data Flow.} \label{fig2}
\end{figure}

Fig. \ref{fig1} illustrates the primary processes in the \textbf{RATAS} framework and Fig. \ref{fig2} highlights its conceptual steps, inputs, and outputs. The framework comprises two main engines, which take an \textit{answer} and a \textit{rubric} as inputs and generate both \textit{whole and partial scores} along with \textit{detailed reasoning} and \textit{score analysis}. These operations are performed by the modules and engines depicted in Fig. \ref{fig3}. All RATAS modules and engines follow an API-based design. Communication between them and with commercial LLM APIs is facilitated by a dedicated \textbf{"LLM Gateway"} module, ensuring seamless integration and interaction across the framework.

\subsection{Converting Rubric to Rubric Knowledge Tree (RKT)}

The RKT is a tree-based structure used to contextualize a rubric, with each node representing a set of scoring rules for specific sections of the rubric. The\textbf{ }root (identified by a rubric ID, (\textit{RID})) corresponds to the entire rubric (Fig. \ref{fig1}), and each child node is a rubric row, referred to as a Rubric Line (\textit{RL}). During each iteration of construction, leaf nodes are expanded into more granular rules and assigned to new child nodes. This \textbf{rule-division} process must satisfy three criteria: (1) The simpler rules must collectively reconstruct the original rule; (2) They must cover distinct, non-overlapping aspects; and (3) Their scoring importance should be approximately equivalent. Depicted by the black arrow in Fig. \ref{fig1}, this expansion continues until no further simplification is possible. The final leaf nodes, representing the fully subdivided rules, are called Simplified Rules\textbf{ }(SR). 
 
\begin{figure}
\includegraphics[width=\textwidth]{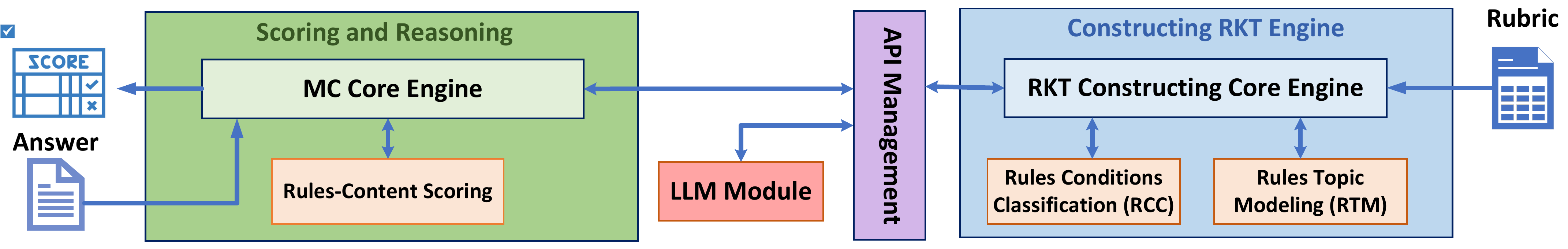}
\caption{RATAS High-Level Architecture.} \label{fig3}
\end{figure}

Each RKT node maintains attributes that capture the characteristics of its scoring rules, divided into: (1) \textbf{contextual attributes}, covering the content and structure of rules, defined during the RKT’s initial creation, and (2) \textbf{scoring attributes}, detailing how rules are assessed, computed during the scoring and reasoning phases. The following outlines each RKT node's attributes, data types, and the RATAS modules and engines (Fig. \ref{fig3}) responsible for their computation. Tree-based structure of RATAS  and the nodes attribute help to generate structured reasoning

Each RKT node holds attributes describing its scoring rules, divided into two categories: 1- \textbf{contextual attributes}, which define the content and structure of each rule at creation, and 2- \textbf{scoring attributes}, computed during the scoring and reasoning phases to assess rule fulfillment. This tree-based architecture, together with these attributes, enables the\textbf{ generation of structured reasoning}. The following sections detail each node’s attributes, data types, and the corresponding RATAS modules (Fig. \ref{fig3}) responsible for their computation.

\begin{itemize}
    \item \textbf{id.}  Each node in the Rubric Knowledge Tree (RKT) is assigned a unique string-based ID by the \textbf{"RKT Constructing Core"}. The root ID is RID, while child IDs append “.i” (an integer) to the parent’s ID, ensuring a hierarchical structure. 

    \item \textbf{leaf.} This boolean attribute indicates whether a node is a leaf (1) or an internal node (0). It is set by “\textbf{core of RKT Constructing}” in each iteration of the RKT construction, allowing the framework to further rule subdivision is possible.

    \item \textbf{criteria.} Represented as a string, this attribute denotes the rule(s) linked to the node, set by the \textbf{"core of RKT Constructing"} module. For child nodes directly under the root, it corresponds to the \textit{Basic-Rule} in the associated rubric row. For deeper nodes, criteria are inherited and further subdivided from the parent’s criteria. 

    \item \textbf{criteria\_simplified\_version.} This attribute is a list of simplified rules, each derived from the node’s main criterion by the \textbf{Criteria Topic Modeling (LLMCTM)} module. Inspired by the NLP topic modeling problem \cite{abdelrazek2023topic}, CTM divides and categorizes the parent’s criteria based on a predefined rule-division policy, ensuring correctness, clarity, and completeness. 

    \item \textbf{separate\_rule\_number.} An integer capturing how many simplified rules are in the criteria\_simplified\_version list. It is periodically updated by the “\textbf{core of RKT Constructing}” module whenever new subdivisions are introduced.

    \item \textbf{score\_source.} A float value representing the Score-Source percentage for the node, set by the \textbf{"core of RKT Constructing"} module. For child nodes under the root, it is initialized from the corresponding rubric row. When rules are divided into finer segments, child nodes inherit proportionally adjusted values to maintain a consistent scoring logic. 

    \item \textbf{score\_source\_ID.} This integer uniquely identifies the score source assigned to each node. Like the id attribute, it is generated by the “\textbf{core of RKT Constructing}” module to preserve traceability, especially in complex rubrics.

    \item \textbf{influence\_on\_scoring.} A float specifying what fraction of the node’s Score-Source is awarded if its rule is satisfied. Computed by the “\textbf{core of RKT Constructing}” engine, this attribute ensures transparent mapping between rule fulfillment and the final score.

    \item \textbf{sub\_condition (quality level).} A list of (string, float) pairs representing potential performance levels—quality levels—and their corresponding scores. For root child nodes, these pairs mirror the rubric’s \textit{Level-of-Achievement}. For deeper nodes, they are generated by the \textbf{LLM “Criteria Sub-condition Classification (LLMCSC)”} module, which functions similarly to text classification \cite{aggarwal2012survey} by classifying a parent node's sub-conditions based on \verb|criteria_simplified_version|, and assigning the corresponding text to each related child node. 

    \item \textbf{Children.} The list of each node's children.
\end{itemize}

\subsection{Scoring Answer based on SRs}

The \textbf{LLM "\textbf{SR-based Scoring and Reasoning (LLMSSR)}"} module computes an answer’s score by determining whether each SR—the smallest indivisible rule—is met (1) or unmet (0). Rather than predicting numeric scores directly, the LLM classifies each SR’s fulfillment, resulting in a more accurate and reliable output. For every SR, this module receives the relevant answer segment and other necessary inputs (Fig. \ref{fig2}). The \textbf{"core scoring and reasoning engine"} then generates text-based explanations based on the corresponding criteria and the matched \textit{la}. As shown by the blue and red dashed arrows in Fig. \ref{fig1}, this granular approach provides a comprehensive analysis of student responses and delivers actionable feedback for both learners and instructors.

\begin{itemize}
    \item \textbf{\textit{score}:} A floating-point value reflecting the overall grade, aggregated from the RKT nodes (for SR nodes, it is equal to 0 or 1).
    \item \textbf{\textit{Reason}:} A structured list or dictionary capturing node-specific details. For leaf nodes, each entry includes the RKT node \textit{id}, \textit{score\_source\_ID}, \textit{influence\_on\_scoring}, \textit{rewarded\_score}, \textit{related\_content}, and \textit{reason\_text}.
\end{itemize}

\subsection{Cascading Score and Reason Propagation}
  
In the concluding steps of RATAS, both \textbf{partial and final scores}—along with their corresponding explanations—are computed. Partial scores capture meaningful combinations of rubric criteria, while final scores reflect the overall assessment. The \textbf{"core of the scoring and reasoning engine"} merges and concatenates unit-level scores and reasons from each SR, propagating them through the RKT (indicated by green dashed arrows in Fig. \ref{fig1}). Key inputs include SR-based scores and reasoning, the RKT structure, and each internal node’s \textit{influence\_on\_scoring}. The engine then updates the \textit{score} and \textit{reason} attributes of all internal nodes, enabling RATAS to provide detailed feedback on answer quality, strengths, and weaknesses. The cascading, tree-based architecture of RATAS enables the generation of structured explanations with different templates that precisely attribute scores to specific rubric components. For example, consider the following partial reasoning template:

\textbf{Answer Analysis:}  For rubric row 1, the answer fully satisfies criteria \textit{"SR1"} and \textit{"SR2"}, achieving the \textit{"corresponding la"} and thus receiving \textit{x1}\% of the \textit{"associated SS"}. In contrast, criterion \textit{"SR3"} is only partially met—earning \textit{x2}\% because of matching with \textit{"corresponding la"}—and criteria \textit{SR3} and \textit{SR4} are not addressed because of the last \textit{"corresponding la"}.

\textbf{Improvement Points:} 1- For partially met criteria: highlight the gap between the maximum achievable \textit{la} and the achieved level. 2- For unmet criteria: Specify the maximum attainable \textit{la}.

\section{RATAS  Algorithms and Methods}

As illustrated in Fig. \textbackslash{}ref\{fig1\}, RATAS consists of two main engines: the \textbf{RKT Constructing Engine}, which analyzes and conceptualizes the rubric, and the \textbf{Scoring and Reasoning Engine}, which generates partial and overall scores along with reasoning outputs. Each engine includes a \textbf{core module} that integrates results from various DNT modules using specialized methods. RATAS is designed to decompose the complex RATG problem into simpler, shorter-input tasks, offering four key advantages: \textbf{improved accuracy}, as simplifying tasks increases the likelihood of high-quality results; and \textbf{manageable} input length, as breaking lengthy inputs into smaller segments enhances LLM output quality and reliability.

\subsection{General responsibilities of Core modules }

Core modules must employ carefully designed models and architectures to divide complex tasks into simpler NLP processes without compromising overall quality. Although each module may use different methods, they share several key responsibilities: \textbf{coordinating} module execution and embedded functions, generating inputs for downstream tasks, \textbf{leveraging} outputs from DNT modules, \textbf{computing} essential results with embedded functions, \textbf{integrating} all intermediate outputs into coherent final results, and \textbf{managing} both internal and external communications to maintain a seamless workflow. Algorithm \ref{alg:RKTAnswerScoring} provides a high-level representation of the two core engines of RATAS.

\begin{algorithm}[ht]
\caption{RKT Construction and Answer Scoring with RKT and ACT}
\label{alg:RKTAnswerScoring}

\textbf{Part 1: RKT Construction}--\textbf{Input:} Rubric Table $T$ of $n$ rows,\ \textbf{Output:} RKT

Create root node $R$, set $R.id=$ Rubric-ID, $R.leaf=1$, $R.criteria=\{\text{all BasicCriteria}\}$, $R.num=n$\\
\For{$x=1$ \KwTo $n$}{
  $R.cs[x] \gets \text{BasicCrit}(x)$;\,
  $R.ans[x] \gets \text{AnsSect}(x)$;\,
  $R.ss[x] \gets \text{ScoreSrc}(x)$;\,
  $R.sc[x] \gets \text{SubCond}(x)$
}
$R.title=\text{"all"};\ R.scoreSrc=\text{"all"};\ R.infl=100\%$\\
\While{\text{any leaf node }N\text{ with }N.leaf=1}{
  \textsc{NodeExpantion}(RKT,\,N)
}

\textbf{Funtion NodeExpantion} (\textbf{Input:} leaf node $N$)\ \textbf{Output: }expanded node
$N.leaf=0$\\
\For{$x=0$ \KwTo $N.num-1$}{
  Create child $M$, $M.id=N.id.x$, $M.leaf=1$\\
  $M.criteria = N.cs[x]$;\,
  $M.cs = \textsc{LLMCTM}(M.criteria)$;\,
  $M.num = |M.cs|$;\,
  $M.infl = \frac{N.infl}{M.num}$\\
  \uIf{$N$ is root child}{
    $M.ans=N.ans[x]$;\,$M.ss=N.ss[x]$;\,$M.ssID=N.ssID[x]$;\,$M.sc=N.sc[x]$
  }
  \Else{
    Inherit $M.ans,\ M.ss,\ M.ssID$ from $N$\\
    $M.sc = \textsc{LLMCSC}(M.cs,\ N.sc[x])$
  }
}

\hrule  % Horizontal line as a clear separator
\vspace{5pt}  % Adjust spacing

\textbf{Part 2: Answer Scoring}--\textbf{Input:} RKT,\ Answer $A$,\ \textbf{Output:} Score,\ Reason for each node

\ForEach{\text{leaf node }N\text{ in RKT}}{
  $[N.nodeScore,\ N.reason[N.id]] \gets \textsc{LLMSSR}(A,\ N.cs,\ N.subCond)$
}

\ForEach{\text{internal node }N\text{ in RKT}}{
  $N.nodeScore \gets \sum_{C \in \text{Children}(N)} (C.infl \times C.nodeScore)$\\
  $N.reason[N.id] \gets \textsc{StructReason}(\{C.reason[C.id] \mid C \in \text{Children}(N)\})$
}
\end{algorithm}

\subsection{Methodology for Addressing DNTs}

One of the primary challenges in RATAS is choosing effective approaches for its downstream tasks since their quality directly affects overall performance. Although these tasks are simpler than RATG, they still qualify as complex NLP problems that some of them suffer from lack extensive domain-specific datasets. Table \ref{tab1} compares key criteria of various LLM-based implementation strategies, guiding the selection of optimal methods for RATAS.

\begin{table}
\caption{Different approaches for employing LLMs in addressing RATAS DNTs}\label{tab2}
\begin{tabular}{|p{6.7cm}|p{5.2cm}|} \hline

\textbf{Approach}&\textbf{Cons}\\ \hline
Fine-tuning Open-Source LLMs (e.g., Falcon \cite{almazrouei2023falcon}, BERT \cite{devlin2018bert}). Pretrained on large corpora; performance scales with parameter count and data size.&Requires extensive data for each domain or exam subject; high computational costs for large-parameter models.\\ \hline
Task-Specific Fine-Tuned LLMs (e.g., Pegasus for Summarization \cite{zhang2020pegasus}). Adapted from various foundation models using large datasets, but typically limited to well-defined tasks.&Existing task-specific LLMs often cannot handle RATAS’s novel downstream NLP tasks. \\ \hline
Open-Source Generative Models (e.g., LLaMA 3.1 \cite{touvron2023llama}, Gemma 2 \cite{team2024gemma}). Pretrained generative architectures adaptable via prompting; flexible but may require substantial computational resources at scale.&Low-parameter models yield poor performance; high-parameter models demand considerable compute and may lack comprehensive domain knowledge.\\ \hline
Black-Box Generative LLM APIs (e.g., Azure OpenAI 4o, BART \cite{lewis2019bart}). Pretrained on vast corpora; produce high-quality, domain-agnostic outputs; well-suited for advanced NLP tasks with minimal local compute.&Token-based usage fees; internal mechanics are opaque. RATAS mitigates this with no need for lower-layer explainability.\\ \hline
\end{tabular}
\end{table}

According to the analysis, given the limitations of ARSN and the DNTs in RATAS, we employ \textbf{GPT-4o}, a powerful black-box LLM API released by OpenAI in May 2024.  According to \cite{ref-url1}, GPT-4o offers faster performance and higher benchmark scores than GPT-4 and GPT-3.5, featuring an expanded context window, improved tokenization, and support for multimodal inputs. Its advanced transformer architecture and higher rate limits (5x those of GPT-4 Turbo) make it well-suited for large-scale AI applications. Furthermore, GPT-4o’s wide knowledge coverage is critical for \textbf{subject-agnostic automated scoring}, enabling RATAS to function without specialized training for each domain \cite{shahriar2024putting}. By relying on GPT APIs, the framework also avoids the \textbf{need for substantial computational resources}. 

Although LLMs can be adapted via fine-tuning, prompt engineering, or retrieval-augmented generation (RAG) \cite{liu2024hift}, \textbf{prompt engineering }is well-suited for RATAS due to its minimal training data requirements. Prompt engineering involves designing and optimizing textual inputs to elicit precise, contextually relevant responses \cite{amatriain2024prompt}. We combine multiple prompting techniques, including Chain-of-Thought (CoT), few-shot examples, and instruction prompting \cite{mishra2021reframing}, to guide GPT-4o in producing accurate and consistent outputs across RATAS’s downstream NLP tasks. We use a crowdsourcing approach to refine prompts, incorporating human evaluation and revision based on feedback generated by AI. For GPT-based tasks, we use the Azure OpenAI 4o Assistant API \cite{MicrosoftOpenAIAssistants}, which integrates advanced conversational capabilities into applications. This setup maintains conversation history, enabling diverse prompting techniques, natural language interactions, and intelligent task automation. The design and implementation aspects of DNTs, including prompts, associated techniques, and example use cases, are documented in the RATAS GitHub repository.

\section{Dataset and Results}

For evaluation, due to the lack of real-world datasets with complex rubrics and lengthy answers, we constructed a unique, contextualized dataset from university-level project-based courses. This dataset comprises 417 selected responses from approximately 1,500 answers, ranging from 14 to 1,285 words (average: 366), significantly exceeding RiceCharm, the existing long-answer datasets with an average of 120 words \cite{sonkar2024automated} . Additionally, it includes authentic, structured rubrics with extensive scoring rules—a complex combination of 34 different criteria (available on the RATAS GitHub). 

To evaluate the performance of RATAS relative to direct GPT-4o for the RATG task, we conducted three runs for each approach. The average outcomes show, \ref{tab5}, that RATAS achieves outstanding accuracy, with a Mean Absolute Error (MAE) of 0.0309, a Root Mean Square Error (RMSE) of 0.0443, and an R² of 0.9627. In comparison, direct GPT-4o yields substantially higher errors (MAE = 0.2355, RMSE = 0.2923), a negative R² of –0.6262, and a low Pearson’s correlation coefficient (r = 0.3743).

To evaluate the impact of response length on model output quality, we divided the dataset into two subsets: responses under 600 words and those exceeding this threshold. When analyzing responses over 600 words, the RATAS metric is slightly lower than its performance on the entire dataset and the subset of shorter responses. However, it still demonstrates strong performance. In contrast, GPT-4o exhibits a more pronounced decline in performance on longer responses compared to RATAS. These findings highlight RATAS’s robustness and consistency across different response lengths, reinforcing its effectiveness over the direct GPT-4o approach. 

We additionally assessed the reliability of both methods by computing the Intraclass Correlation Coefficient (ICC) across the full dataset. RATAS achieves a notably higher ICC \textbf{(0.9662)} than direct GPT-4o \textbf{(0.5984)}, indicating superior reliability. This improvement stems from RATAS’s architecture, which leverages simpler and more focused inputs (DNTs), as opposed to applying direct GPT-4o to the more complex RATG problem.
 
\begin{table}
\caption{Evaluation Metrics for RATAS and GPT-4o on Whole Dataset and Subsets of Answers }\label{tab5}
\centering
\begin{tabular}{|p{1.1cm}|p{4cm}|>{\centering\arraybackslash}p{1.6cm}|>{\centering\arraybackslash}p{1.6cm}|>{\centering\arraybackslash}p{1.6cm}|>{\centering\arraybackslash}p{1.6cm}|} 
\hline
\textbf{Method} & \textbf{Dataset} & \textbf{MAE} & \textbf{RMSE} & \textbf{R²} & \textbf{Pearson’s r} \\ 
\hline
RATAS & Whole dataset & 0.0309 & 0.0443 & 0.9627 & 0.9813 \\ 
\hline
RATAS & Dataset related to answers with fewer than 600 words & 0.0280 & 0.0387 & 0.9696 & 0.9848 \\ 
\hline
RATAS & Dataset related to answers with more than 600 words & 0.0603 & 0.0810 & 0.8874 & 0.9519 \\ 
\hline
GPT 4o & Whole dataset & 0.2355 & 0.2923 & -0.6262 & 0.3743 \\ 
\hline
GPT 4o & Dataset related to answers with fewer than 600 words & 0.2313 & 0.2878 & -0.6793 & 0.4227 \\ 
\hline
GPT 4o & Dataset related to answers with more than 600 words & 0.2768 & 0.3345 & -0.9181 & -0.2582 \\ 
\hline
\end{tabular}
\end{table}

%relibility of using LLMs
\section{Conclusion}
  
 In this research, we introduce \textbf{RATAS}, a novel framework for \textbf{Rubric-Based Automatic Text Grading (RATG)} designed to handle real-world open-ended (OE) exams with extensive rubrics. This framework meets various stakeholder requirements, including reliable scoring, support for diverse exam formats, and adherence to authentic rubric-based evaluation. Our empirical findings highlight RATAS’s strong performance, addressing limitations of existing methods through precise floating-point scoring, structured feedback generation, and the capacity to evaluate very long answers beyond the existing ASAG and ALAG datasets. 

Although RATAS demonstrates effectiveness across various answer lengths, its performance declines for very long responses (exceeding 600 words). Future work should focus on optimizing RATAS for such extensive inputs and developing specialized frameworks for lengthy academic reports (20–50 pages). Moreover, performance may vary under more intricate rubrics featuring hierarchical or interdependent scoring criteria, which require further evaluation through the creation of dedicated datasets. 


%\printbibliography
\bibliography{references}   % use the actual name of your .bib file

\end{document}